\documentclass[11pt]{article}

\usepackage[preprint]{acl}

\usepackage{times}
\usepackage{latexsym}

\usepackage[T1]{fontenc}
\usepackage[utf8]{inputenc}

\usepackage{microtype}

\usepackage{inconsolata}

\usepackage{graphicx}

\usepackage{booktabs}
\usepackage{tabularx}
\usepackage{array}
\usepackage{placeins}
\usepackage{needspace}
\usepackage{float}
\usepackage{threeparttable}
\usepackage[most]{tcolorbox}
\usepackage{xcolor}
\usepackage{booktabs}
\usepackage{url} % or hyperref
\usepackage{tabularx}
\usepackage[table]{xcolor}
\usepackage[most]{tcolorbox}
\usepackage{capt-of}
\usepackage{listings}
\usepackage{enumitem}
\usepackage{fvextra}

\usepackage{color}

\lstdefinelanguage{PLSQL}{
  language=SQL,
  morekeywords={BEGIN, END, DECLARE, DBMS_OUTPUT, PUT_LINE},
  sensitive=false
}

\title{\textsc{PLSQLBench}: Benchmarking LLM Systems for Executable Procedural Database Programming}

\author{
  \textbf{Marianne Menglin Liu\textsuperscript{1,*}},
  \textbf{Leonid Boytsov\textsuperscript{1,*}},
  \textbf{Daniel W. Peterson\textsuperscript{1}},
  \textbf{Pramuditha Perera\textsuperscript{1}},
\\
  \textbf{Rongguang Wang\textsuperscript{1}},
  \textbf{Sai Ashish Somayajula\textsuperscript{1}},
  \textbf{Syed Hamza Rafique\textsuperscript{2}},
  \textbf{Rohit Saini\textsuperscript{2}},
\\
  \textbf{Shubham Pathak\textsuperscript{2}},
  \textbf{Sujeeth Bharadwaj\textsuperscript{1}},
  \textbf{Tao Sheng\textsuperscript{1}},
  \textbf{Graham Horwood\textsuperscript{1}},
\\
  \textbf{Fahad Shah\textsuperscript{1}},
  \textbf{Ankan Bansal\textsuperscript{1}},
  \textbf{Sujith Ravi\textsuperscript{1}},
  \textbf{Dan Roth\textsuperscript{1}}
\\
\\
  \textsuperscript{1}Oracle AI,
  \textsuperscript{2}Turing Enterprise Inc
\\
\small{
  \textsuperscript{*}Equal contribution.
  \textbf{Correspondence:}
  \{marianne.liu, leo.boytsov, dan.roth\}@oracle.com
}
}

\begin{document}
\maketitle
\begin{abstract}

We present \textsc{PLSQLBench}, to our knowledge the first benchmark for evaluating whether LLMs can
write executable PL/SQL programs, with correctness measured through
execution-based tests. Existing LLM evaluations largely target general-purpose
code generation or declarative text-to-SQL, leaving procedural database
programming underexplored. \textsc{PLSQLBench} contains 2,865 instances:
2,594 single-turn tasks and 271 multi-turn conversations spanning 978 turns.
The benchmark combines complex schema-grounded tasks over enterprise-style
Spider~2 databases, simpler schema-grounded tasks derived from Spider, and
MBPP-derived procedural problems, covering varying levels of database grounding
and procedural complexity. Experiments with eight
LLMs reveal recurring difficulties in schema grounding, PL/SQL dialect
fidelity, procedural control flow, exception handling, and cross-turn
consistency. Tool-augmented LLM agents improve performance on several
schema-grounded evaluations, although substantial gaps remain. These results
highlight procedural database-programming capabilities not directly assessed
by conventional code-generation or text-to-SQL benchmarks. Our code is available
at \url{https://github.com/oracle-samples/plsqlbench}.

\end{abstract}

\begin{figure*}[t]
  \centering
  \includegraphics[width=0.9\textwidth]{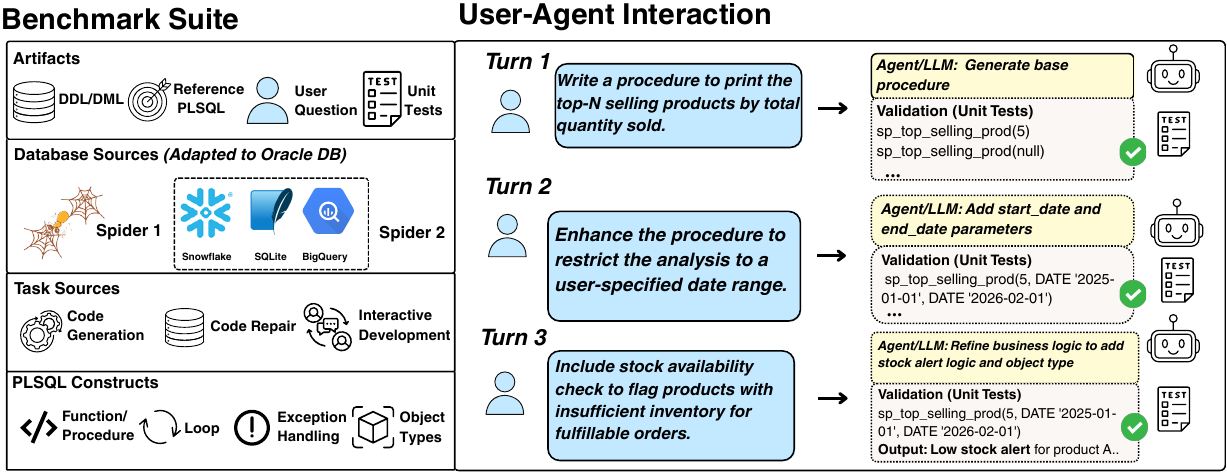}
\caption{Overview of \textsc{PLSQLBench}. \textsc{PLSQLBench} combines PL/SQL artifacts, database schemas, development
tasks, and executable tests to evaluate whether LLM systems can generate,
modify, debug, and repair procedural programs while preserving conversational
context. It contains both single-turn tasks and multi-turn interactive workflows
requiring generation, modification, debugging, validation, and repair of
executable PL/SQL programs.}
  \label{fig:archi}
\end{figure*}

\section{Introduction}
Enterprise database development often requires procedural database programming:
creating reusable database-resident programs that combine SQL with procedural
constructs such as variables, control flow, loops, cursors, and error handling. 
In production systems, these programs validate records, enforce
business rules, log failures, coordinate transactions, and run scheduled
maintenance tasks.
Procedural programming over databases is supported across relational and
cloud data platforms through procedural SQL dialects, including GoogleSQL
scripting in BigQuery, PL/pgSQL in PostgreSQL, Snowflake Scripting in Snowflake,
Transact-SQL in SQL Server, and PL/SQL in Oracle Database
\citep{googlecloud_bigquery_procedural_language,postgresql_plpgsql,
snowflake_scripting_guide,microsoft_tsql_reference,oracle_plsql_developers}.

However, current benchmarks for AI-assisted database development remain centered
on either general-purpose programming or standalone declarative SQL. General code-generation benchmarks such as HumanEval, MBPP, MBPP+,
and MultiPL-E primarily evaluate programming in languages such as Python, Java,
and C++ \citep{chen2021evaluating,austin2021program,cassano2022multipl,liu2024evaluating}.
Text-to-SQL benchmarks such as Spider, BIRD, and Spider 2.0 focus on generating
standalone declarative queries over relational schemas
\citep{yu2018spider,li2023can,lei2025spider}. Even recent interactive database
benchmarks such as Bird-Interact \citep{huo2025bird} remain centered on SQL
querying rather than executable procedural database programs. As a result, the
ability of AI systems to generate, modify, debug, and repair procedural database
programs remains underexplored.

To study this setting concretely, we introduce \textsc{PLSQLBench}, a benchmark
for evaluating LLM systems on executable procedural database programming in
PL/SQL. 
In this work, we focus on PL/SQL, an executable procedural SQL
dialect with support for stored procedures, functions, packages, cursors and 
exceptions. Oracle Database is currently ranked
first in the DB-Engines popularity ranking of database management systems
\citep{dbengines_ranking}, making PL/SQL a practically relevant testbed for
evaluating AI systems on executable procedural database programs. Figure~\ref{fig:archi} illustrates the benchmark
design through a representative workflow: a user asks a system to write a
reusable procedure for reporting top-selling products, then iteratively extends
it with date-range filtering and low-stock alerts. \textsc{PLSQLBench} combines PL/SQL artifacts, database schemas, development
tasks, and executable tests to evaluate whether LLM systems can generate,
modify, debug, and repair procedural programs while preserving conversational
context.

Our contributions are threefold:
\begin{enumerate}
    \item We introduce \textsc{PLSQLBench}, to our knowledge the first benchmark
specifically designed to evaluate LLM systems on executable procedural database
programming in PL/SQL. It combines complex schema-grounded tasks over
enterprise-style Spider~2 databases, simpler schema-grounded tasks derived from
Spider, and MBPP-derived procedural problems, covering varying levels of
database grounding and procedural complexity. Unlike conventional text-to-SQL
benchmarks that emphasize standalone declarative queries, \textsc{PLSQLBench}
evaluates executable procedural programs involving control flow, schema-grounded
logic, and exception handling, in both single-turn and multi-turn development
settings.
\item We design an execution-based evaluation protocol for procedural
database workflows, combining strict and partial-credit metrics for
single-turn tasks with dynamic multi-turn evaluation under accumulated
conversation context.
\item We benchmark eight proprietary and open-weight LLMs, together with
tool-augmented database agents, revealing persistent challenges in schema use, PL/SQL dialect fidelity, procedural reasoning, exception handling, and
cross-turn consistency.
\end{enumerate}

\section{Related Work}
\paragraph{Code-generation benchmarks.}
General code-generation benchmarks such as HumanEval and MBPP evaluate whether
models can synthesize short programs from natural-language specifications, often
using unit tests for execution-based scoring
\citep{chen2021evaluating,austin2021program}. MultiPL-E and MXEVAL extend this evaluation
across multiple programming languages \citep{cassano2022multipl,DBLP:conf/iclr/AthiwaratkunGWL23}. Recent benchmarks
broaden coverage to contamination-resistant evaluation, complex library use, and
repository-level software engineering through LiveCodeBench, BigCodeBench, and
SWE-bench \citep{jain2025livecodebench,zhuo2025bigcodebench,jimenez2024swe}.
However, these benchmarks focus on general-purpose coding tasks, not executable
database programs whose correctness depends on schemas, persistent objects,
transactions, and procedural runtime behavior.

\paragraph{Text-to-SQL and database interaction.}
Text-to-SQL benchmarks have evolved from cross-domain query generation to more
realistic enterprise database workflows. Spider evaluates complex
schema-grounded SQL generation, while BIRD, BEAVER, and Spider 2.0 scale this
setting to larger databases, enterprise-style schemas, and workflow-level tasks
such as Spider2-DBT
\citep{yu2018spider,li2023can,chen2024beaver,lei2025spider}. Conversational and
interactive benchmarks, including SParC, CoSQL, and BIRD-INTERACT, study
context-dependent querying, clarification, and execution feedback
\citep{yu2019sparc,yu2019cosql,huo2025bird}, and BIRD-CRITIC targets SQL issue
debugging \citep{li2026swe}. Despite this progress, these benchmarks remain
centered on SQL-based tasks. \textsc{PLSQLBench} instead evaluates procedural database programming, 
where models must write executable PL/SQL programs that combine procedural logic with SQL over database schemas.

\paragraph{Enterprise workflow benchmarks.}
Enterprise-oriented benchmarks increasingly move beyond isolated prompts to
realistic agent settings, including web-based workplace tasks, simulated
software-company environments, and retrieval over heterogeneous enterprise
artifacts \citep{boisvert2024workarena++,xu2026theagentcompany,
choubey2025benchmarking}. 
% ML version
% These benchmarks capture enterprise interaction and
% information access, but do not target procedural database development.
% \textsc{PLSQLBench} fills this gap by combining single-turn PLSQL generation
% with multi-turn workflows for modifying, debugging, and repairing procedural
% database programs.
These benchmarks capture enterprise interaction and information access, but they do not evaluate executable procedural database programming. \textsc{PLSQLBench} addresses this gap by evaluating read-only PL/SQL generation in both single-turn and multi-turn settings. Its tasks span complex schema-grounded scenarios over enterprise-style
databases, simpler schema-grounded database tasks, and MBPP-derived procedural
problems, enabling evaluation across varying levels of database grounding and
procedural complexity.

\section{Dataset Construction and Curation}
\label{sec:data-generation}

\subsection{Benchmark Composition}
We construct \textsc{PLSQLBench} from three source families: Spider 2.0 Lite,
Spider 1.0, and MBPP/MBPP+, forming five subsets: Spider2-ST, Spider2-MT,
Spider-PLSQL, MBPP-PLSQL, and MBPP+-PLSQL. Spider2-ST and Spider2-MT contain
new PL/SQL tasks curated over Oracle-normalized Spider 2.0 Lite schemas;
Spider-PLSQL retains the original Spider questions and uses gold-SQL result
sets for execution-based evaluation; and MBPP-PLSQL and MBPP+-PLSQL adapt
short programming problems and their tests to PL/SQL. These subsets
cover varying levels of database grounding, schema complexity, and procedural
reasoning.

The \textsc{PLSQLBench} construction pipeline comprises source normalization
and schema setup, question generation and prompt construction, reference
PL/SQL curation, unit-test generation, and quality control. Dataset-specific
construction and evaluation procedures are described in the corresponding
appendices, with representative examples in
Appendix~\ref{app:data-example}.

Overall, \textsc{PLSQLBench} contains 2,865 instances: 2,594 single-turn tasks
and 271 multi-turn conversations spanning 978 turns.
Table~\ref{tab:dataset-stats} summarizes the subset-level statistics.
For safe and repeatable execution, benchmark tasks do not require modifying
persistent data through inserts, updates, or deletes, avoiding destructive
side effects across runs.

Spider2-ST and Spider2-MT are the primary enterprise-style, schema-grounded
subsets, using Spider 2.0 Lite databases ported to the Oracle dialect
(see Appendix~\ref{app:spider2-oracle-normalization}).
Spider2-ST contains 407 development and 103 test tasks. Spider2-MT contains
208 development and 63 test complete conversations, spanning 750 and 228
turns, respectively. Detailed turn, difficulty, and reasoning-type
distributions are provided in Appendix~\ref{app:data_stats_details}.

Spider-PLSQL contains 970 tasks over Spider 1.0 databases ported to the Oracle
dialect \citep{spider_oracle_conversion}. MBPP-PLSQL and MBPP+-PLSQL contain
806 and 308 tasks, respectively. These subsets adapt the original programming
problems and Python unit tests into executable PL/SQL function-generation
tasks. They do not require persistent database tables, although some tasks
require Oracle object or collection types to represent structured inputs,
providing a complementary evaluation of procedural reasoning with less
database grounding than the Spider-derived subsets.

\subsection{Source Normalization and Schema Setup}
\label{sec:schema-setup}

For schema-grounded tasks, each prompt includes the database context needed to
write and execute the requested PL/SQL program, including table definitions,
column names, types, and any additional objects referenced by the task. Across
the benchmark, source tasks are adapted differently depending on the dataset:
MBPP and MBPP+ tasks are mapped to PL/SQL functions or procedures that do not
require persistent database tables, with Oracle types provided when needed; Spider 1.0 retains its original
natural-language questions and Oracle schemas while prompting for PL/SQL with
SQL-output-compatible formatting; and Spider 2.0 Lite tasks are curated
directly over Oracle-dialect schemas prepared for PL/SQL evaluation. The
dataset-specific normalization and curation procedures are described in
Appendices~\ref{app:spider2-oracle-normalization},
\ref{app:spider1-plsql-conversion}, and
\ref{app:mbpp-plsql-conversion}.

%\subsection{Question Generation}
\subsection{Spider 2.0 Question Generation and Prompt Construction}
\label{sec:question-generation}

For Spider2-ST and Spider2-MT, we construct natural-language prompts for three task families: PL/SQL code
generation, schema-grounded code repair, and interactive development. Code
generation asks models to synthesize new PL/SQL artifacts from a specification;
code repair asks them to correct faulty or incomplete code with respect to the
schema and expected behavior; and interactive development uses multi-turn
requests that extend, modify, debug, or repair earlier code while preserving
prior context. Prompt construction follows the metadata-driven curation protocol
in Appendix~\ref{app:task-curation-methodology}. The system prompt used for generation can be found in Appendix \ref{app:prompts}. 

% Data statistics table
\begin{table}[t]
\centering
\small
\begin{threeparttable}
\setlength{\tabcolsep}{3pt}
\renewcommand{\arraystretch}{1.08}
\begin{tabular}{lrrr}
\toprule
\textbf{Benchmark}
& \textbf{Inst.}
& \textbf{\#DB}
& \textbf{\#Test (Avg.)} \\
\midrule

Spider2-ST-dev
& 407 & 26 & 4.9 \\

Spider2-ST-test
& 103 & 8 & 4.7 \\

Spider2-MT-dev
& 208 & 26 & 13.7 \\

Spider2-MT-test
& 63 & 7 & 13.0 \\

Spider-PLSQL
& 970 & 20 & 1.0 \\

MBPP-PLSQL$^\dagger$
& 806 & -- & 1.0 \\

MBPP+-PLSQL$^\dagger$
& 308 & -- & 1.0 \\

\midrule
\textbf{PLSQLBench (Overall)}
& \textbf{2,865}
& \textbf{54}
& \textbf{2.9} \\

\bottomrule
\end{tabular}

\begin{tablenotes}[flushleft]
\footnotesize
\item[$^\dagger$] MBPP-derived function-generation tasks do not require
persistent database tables, although some use Oracle object or collection
types.
\item[] For Spider2-MT, an instance denotes a complete conversation, and the test
average counts executable tests across all turns.
\end{tablenotes}

\caption{
Dataset statistics for \textsc{PLSQLBench}. Inst.\ denotes the number of
single-turn tasks or multi-turn conversations, \#DB the number of distinct
databases, and \#Test (Avg.) the average number of executable unit tests per
instance.
}
\label{tab:dataset-stats}
\end{threeparttable}
\end{table}

% Main base-model results
% Excludes Codex DB Agent results and Spider-SQL

\begin{table*}[t]
\centering
\small
\setlength{\tabcolsep}{3.5pt}
\renewcommand{\arraystretch}{1.08}
\begin{tabular}{
  @{}l
  *{4}{>{\hspace{0.5pt}}c<{\hspace{0.5pt}}}|
  *{3}{>{\hspace{0.5pt}}c<{\hspace{0.5pt}}}|
  c@{}
}
\toprule
& \multicolumn{4}{c|}{\textbf{Development}}
& \multicolumn{3}{c|}{\textbf{Test}}
& \textbf{Overall} \\
\cmidrule(lr){2-5}
\cmidrule(lr){6-8}
\cmidrule(l){9-9}
\textbf{Model}
& MBPP
& \shortstack{Spider-\\PLSQL}
& \shortstack{Spider2-\\ST}
& \shortstack{Spider2-\\MT}
& MBPP+
& \shortstack{Spider2-\\ST}
& \shortstack{Spider2-\\MT}
& \shortstack{Test\\Mean} \\
\midrule

\rowcolor{gray!15}
\multicolumn{9}{@{}l}{\textit{Open-weight models}} \\
Llama-4-Maverick & 57.82 & 75.46 & 60.10 & 48.38 & 26.30 & 61.24 & 52.85 & 46.80 \\
Gemma-4-31B & 80.40 & 81.44 & 73.90 & 67.90 & 39.94 & 76.26 & 70.57 & 62.26 \\

\rowcolor{gray!15}
\multicolumn{9}{@{}l}{\textit{Proprietary models}} \\
Gemini-2.5-Flash-Lite & 63.77 & 73.92 & 60.99 & 50.16 & 33.44 & 60.94 & 56.78 & 50.39 \\
Grok-4.3 & 87.59 & 64.95 & 67.28 & 63.45 & 38.31 & 66.98 & 65.46 & 56.92 \\
GPT-5.4-Mini & 70.35 & 76.60 & 63.14 & 61.29 & 38.96 & 70.90 & 65.94 & 58.60 \\
GPT-5.4 & 83.75 & 78.45 & \textbf{74.35} & \textbf{69.08} & 42.21 & \textbf{79.47} & \textbf{73.19} & \textbf{64.96} \\
GPT-5.6-Sol & \textbf{96.28} & 79.59 & 61.77 & 64.69 & 46.75 & 71.57 & 66.81 & 61.71 \\
Claude-Opus-4.8 & 90.32 & \textbf{84.33} & 73.92 & 65.52 & \textbf{47.73} & 76.34 & 66.40 & 63.49 \\
\bottomrule
\end{tabular}
\caption{
Main \textsc{PLSQLBench} results, with overall means computed only across
the three test sets. All values are Mean Test Pass@1 (\%).
Best results in each column are shown in bold.
}
\label{tab:main-results}
\end{table*}

\subsection{Execution-Based Evaluation}
\label{sec:test-construction}
For Spider2-ST and Spider2-MT, each task contains an executable reference
PL/SQL answer and one or more test scripts, but the expected test outputs are
not stored in advance. We first execute the tests against the reference answer
and capture the resulting outputs. We then execute the same tests against the
generated answer and compare its outputs with the captured reference outputs.
For Spider2-MT, we use rollout evaluation: reference and generated answers are
executed in turn order in separate clean states, so each later turn observes
the state established by earlier turns in the same sequence, without mixing
generated and reference answers.

For MBPP-PLSQL and MBPP+-PLSQL, we retain the original programming problems
and ``translate'' their Python unit tests into self-contained executable PL/SQL
test blocks (using a rule-based approach). 
The translated tests encode the test inputs and expected outputs
directly in PL/SQL and execute entirely in Oracle against the generated PL/SQL
function. No Python code or canonical Python solution is executed during
evaluation. The rule-based unit-test conversion and its validation are
described in Appendix~\ref{app:mbpp-plsql-conversion}.

For Spider-PLSQL, we execute the original Spider 1.0 gold SQL query and use its
result set as the expected output. 
The model is instructed to emit one line per
row, with field values separated by commas, so that the generated PL/SQL output
can be compared directly with the gold SQL result. 
Thus, Spider-PLSQL does not
use reference PL/SQL solutions. Appendix~\ref{app:prompts} provides a sample
Spider-PLSQL prompt illustrating this output-format requirement.
Because generated PL/SQL may emit fields in
a different order from the columns selected by the gold query, we compare its
output against a bounded set of permutations of the gold result-set columns.

\subsection{Quality Control}
\label{sec:quality-control}
We apply multi-stage quality control to Spider2-ST and Spider2-MT to ensure
that each curated example is clear, schema-consistent, executable, and properly
annotated. Each instance is checked for prompt-answer alignment, table and
column accuracy, PL/SQL syntax and execution correctness, test-case validity,
metadata consistency, and required PL/SQL feature coverage. For multi-turn
examples, we additionally verify that later turns preserve prior intent while
introducing realistic incremental changes. Examples with ambiguous prompts,
schema mismatches, non-executable reference code, invalid tests, incomplete
metadata, or inconsistent cross-turn state are revised or removed. Further
details of the review pipeline and quality dimensions are provided in
Appendix~\ref{app:quality-control}. 

The separate quality-control and validation procedures for MBPP-PLSQL and
MBPP+-PLSQL are described in Appendix~\ref{app:mbpp-plsql-conversion}.

\section{Experiments}
\subsection{Experimental Setting}

\textbf{Model and Hardware Setup.}
We evaluate eight proprietary and open-weight models:
Gemma-4-31B, Llama-4-Maverick, Claude-Opus-4.8, Gemini-2.5-Flash-Lite,
Grok-4.3, GPT-5.4-Mini, GPT-5.4, and GPT-5.6-Sol.
We also evaluate two Codex CLI-based
agentic database agents with GPT-5.4-Mini
and GPT-5.6-Sol backbones, using Oracle SQL/PLSQL skills
\citep{oracle_plsql_skills,oracle_sql_dev_skills} and database tool.
Prompts, hyperparameters, and hardware details are provided in
Appendices~\ref{app:prompts} and~\ref{app:hyperparams}.

\textbf{Data Split.}
We use database-disjoint development and private test splits for Spider2-ST
and Spider2-MT, assigning all tasks from the same database to the same split.
MBPP+-PLSQL and the Spider2 test splits are evaluation sets. The Spider2 test splits are fully held out and database-disjoint from development data. MBPP+-PLSQL shares problem descriptions with MBPP-PLSQL but uses expanded unit tests; we withhold the translated PL/SQL unit tests and use them only for evaluation.

\subsection{Evaluation Metrics}

We evaluate generated PL/SQL using execution-based unit tests, following prior
code-generation benchmarks \citep{chen2021evaluating,austin2021program}. Since
a task may contain multiple unit tests, we report two complementary metrics.
\emph{Mean Test Pass@1} is our main metric. It gives partial credit: For each task, we compute the
fraction of its unit tests passed by the model's single generated solution, and
then average this fraction across tasks. \emph{Suite Pass@1} is stricter: a task
is counted as correct only if all of its unit tests pass.

For multi-turn tasks, we carry the previous user requests and
model-generated PL/SQL responses across turns, so that each turn is
conditioned on the accumulated conversation history. We report
Turn Suite Pass@1, which applies suite-level correctness at each turn,
and Episode Pass@1, which requires all turns in a conversation to be
solved. Formal metric definitions are provided in Appendix~\ref{app:metrics}.

\subsection{Experimental Results}
\label{sec:experimental_results}

Table~\ref{tab:main-results} reports the main \textsc{PLSQLBench} results using
Mean Test Pass@1. Appendix~\ref{app:additional_results} reports stricter
execution-based metrics. We highlight the following findings.

\textbf{Frontier proprietary models perform best overall, while strong
open-weight models remain competitive.}
Table~\ref{tab:main-results} shows that GPT-5.4 achieves the highest
test mean Pass@1 at 64.96\%, followed by Claude-Opus-4.8 at 63.49\%.
The relatively small 1.47-point gap between the top two models suggests
that performance is competitive even among frontier systems.
Gemma-4-31B is the strongest open-weight model, ranking third overall
with 62.26\% and outperforming GPT-5.6-Sol at 61.71\%. Performance also
varies substantially by task family: GPT-5.6-Sol leads on MBPP,
Claude-Opus-4.8 on Spider-PLSQL, and GPT-5.4 on both Spider2 variants.
Thus, no single model dominates across all procedural database settings,
and even the best model remains below 65\% mean Pass@1 across the test
sets, indicating substantial headroom in procedural database programming.

\textbf{Tool-augmented agents substantially improve database-grounded
PL/SQL generation.}
Across Tables~\ref{tab:codex-agent-comparison},
\ref{tab:single-turn-suite}, and~\ref{tab:mt-strict-metrics}, Codex DB
Agent improves every Spider2 result for both backbones. On the test sets,
Mean Test Pass@1 increases by 6.9--9.8 points for GPT-5.4-Mini and
12.0--14.5 points for GPT-5.6-Sol, with GPT-5.6-Sol showing consistently larger gains. The improvements also persist under
strict correctness: Spider2-ST Suite Pass@1 increases by 1.9 and 9.7
points, while Spider2-MT Episode Pass@1 increases by 4.8 and 19.1 points.
The largest gain is observed for GPT-5.6-Sol on multi-turn episodes,
where database interaction helps maintain correctness across successive
modifications. In contrast, MBPP+ changes by $+6.2$ and $-0.7$ points,
suggesting that the benefits of database interaction are primarily concentrated
in schema-grounded tasks rather than less database-grounded procedural
programming.
\textbf{Strict correctness remains limited, especially in multi-turn
settings.}
Table~\ref{tab:single-turn-suite} reports Suite Pass@1, which requires
all unit tests for a task to pass. Among direct-generation models,
GPT-5.6-Sol achieves the highest single-turn test mean at 55.90\%, while
GPT-5.4 leads Spider2-ST at 68.93\%. Tool augmentation raises the best
test mean to 60.43\% with GPT-5.6-Sol, but still leaves a substantial fraction of
complete tasks unsolved. The bottleneck is larger in multi-turn settings
(Table~\ref{tab:mt-strict-metrics}), where GPT-5.4 achieves 33.33\%
Episode Pass@1 despite reaching 59.21\% Turn Suite Pass@1. The strongest
tool-augmented agent improves these scores to 41.27\% and 64.91\%,
respectively, but still completes fewer than half of the conversations
without error. These gaps show that errors compound across successive
modifications, making end-to-end multi-turn PL/SQL development and repair
particularly challenging.

\subsection{Error Analysis}
\label{sec:error_analysis}
We analyze \textsc{PLSQLBench} failures across three capabilities: implementing
the requested behavior, generating valid PL/SQL, and preserving the expected
database interface. Appendices~\ref{app:error_analysis_protocol}
and~\ref{app:error_examples} describe the annotation protocol and provide
representative examples.

\textbf{Wrong procedure logic or output.}
The largest failure class (11,015; 82.5\%) consists of procedures that compile
or partially execute but produce incorrect results, including 9,953 output
mismatches and 1,062 runtime failures. Common errors involve filters, lookup
keys, joins, ordering, aggregation, formatting, and edge cases. In MBPP-style
tasks, models often miss boundary cases or return incorrectly formatted values;
in schema-grounded tasks, they may retrieve the wrong rows, aggregate at the
wrong level, or return values in the wrong order. Thus, syntactically plausible
procedures frequently fail to implement the intended semantics.

\textbf{Invalid or incomplete PL/SQL artifacts.}
The second largest category consists of invalid or incomplete generated code,
covering 1,777 failures (13.3\%). This includes 943 syntax or parser failures
and 834 invalid compiled objects that fail when invoked by the test harness.
Typical problems include incomplete procedure bodies, unclosed SQL or PL/SQL
blocks, incorrect declarations, and non-executable explanatory text. These
failures prevent meaningful semantic comparison and show that reliable PL/SQL
artifact construction remains challenging.

\textbf{Interface, signature, or schema grounding errors.}
A smaller but important class of failures involves mismatches between the
generated artifact and the expected database interface. These account for 566
failures (4.2\%). Typical errors include missing expected entry points, wrong
parameter signatures, missing package members, invalid object names, and
incorrect column or identifier references. Spider2-MT accounts for 525 of these
566 failures (92.8\%), suggesting that multi-turn settings make it harder for
models to preserve interface contracts and schema grounding across edits.

\section{Conclusion}
\label{sec:conclusion}

We presented \textsc{PLSQLBench}, to our knowledge the first benchmark
specifically designed to evaluate LLM systems on executable procedural database
programming in PL/SQL, across varying levels of database grounding and
procedural complexity. By combining
schema-grounded and procedural tasks with execution-based tests and
single-turn and multi-turn workflows, \textsc{PLSQLBench} evaluates
capabilities not directly assessed by conventional code-generation or
text-to-SQL benchmarks. Our experiments show that procedural database
programming remains challenging: the best overall Mean Test Pass@1 is 64.96\%,
with substantial drops under stricter metrics. Error analysis reveals
persistent failures in procedure semantics, PL/SQL artifact construction, and
interface or schema grounding, highlighting the need for LLMs that reason more
reliably over executable database semantics, dialect constraints, and
iterative database development.

\section*{Limitations}

\textsc{PLSQLBench} covers only a subset of real-world database development.
Production systems often involve larger schemas, legacy dependencies,
performance constraints, permission boundaries, and deployment requirements that
are not fully captured by the benchmark.

\textsc{PLSQLBench} currently supports only read-only PL/SQL programs. This omits important write-oriented workflows, but keeps execution-based evaluation reproducible by running each prediction against the same fixed database state. Extending the benchmark to updatable PL/SQL is challenging because DML may be interleaved with DDL, commits, rollbacks, and session-level side effects that are difficult to isolate or undo. We leave this direction to future work.

As with any execution-based benchmark, correctness is bounded by test coverage:
passing all tests does not guarantee that a program is correct under all inputs.

\textsc{PLSQLBench} does not provide a large benchmark-specific training set and
is intended for test-only evaluation. Thus, results should be interpreted as a
measure of models' ability to generalize to its schemas, task formats, and test
cases, rather than their ability to adapt through benchmark-specific tuning.

\section*{Ethical Considerations}

\textsc{PLSQLBench} is intended to evaluate the reliability of language models
for procedural database programming, not to encourage unsupervised deployment of
model-generated database code. Because PL/SQL programs can encode business
logic and interact with database state, incorrect generations may lead to
wrong reports, broken interfaces, or unsafe operational behavior if used
directly in production. We therefore evaluate models in isolated benchmark
schemas with database-backed tests and focus on read-only tasks that avoid
persistent data modifications. The benchmark is constructed from public,
curated, or synthetic sources and does not include private customer data or
production database contents. Some tasks and prompts are produced or converted
with LLM assistance, which may introduce artifacts or distributional biases; we
mitigate this through human review, schema-consistency checks, compilation and
execution validation, and test-based quality control. Results should be used to
understand model limitations and guide safer AI-assisted database development,
with human review, access controls, and production testing remaining necessary
for real deployments.

\bibliography{custom}

\newpage
\appendix

% Direct generation vs. Codex DB Agent
\section{Direct Generation vs.\ Tool-Augmented Agent}
\label{app:agentic-comparison}
\begin{table}[H]
\centering
\scriptsize
\setlength{\tabcolsep}{1.7pt}
\renewcommand{\arraystretch}{1.03}
\begin{tabular}{@{}llcc|ccc@{}}
\toprule
& & \multicolumn{2}{c|}{\textbf{Development}}
& \multicolumn{3}{c}{\textbf{Test}} \\
\cmidrule(lr){3-4}
\cmidrule(l){5-7}
\textbf{Backbone} & \textbf{Setting}
& \textbf{S2-ST} & \textbf{S2-MT}
& \textbf{MBPP+} & \textbf{S2-ST} & \textbf{S2-MT} \\
\midrule

GPT-5.4 Mini & Direct & 63.14 & 61.29 & 38.96 & 70.90 & 65.94 \\
& Codex Agent & \textbf{75.36} & \textbf{70.21} & \textbf{45.13} & \textbf{77.80} & \textbf{75.72} \\

\addlinespace[1pt]

GPT-5.6-Sol & Direct & 61.77 & 64.69 & \textbf{46.75} & 71.57 & 66.81 \\
& Codex Agent & \textbf{77.53} & \textbf{73.94} & 46.10 & \textbf{83.56} & \textbf{81.29} \\
\bottomrule
\end{tabular}
\caption{
Direct generation vs.\ Codex DB Agent with medium reasoning.
Results are Mean Test Pass@1 (\%); bold indicates the better setting
for each backbone. MBPP+ is evaluated on its test split. S2-ST and S2-MT
denote Spider2 single- and multi-turn.
}
\label{tab:codex-agent-comparison}
\end{table}

\section{Details on Data Curation and Quality Control}
\label{app:quality-control}

\subsection{Metadata-Guided Curation}
\label{app:metadata-guided-curation}

Each PL/SQL task is initialized with prepopulated metadata specifying the source
dataset, database, target PL/SQL constructs, reasoning types, difficulty level,
and number of turns. Developers are instructed to follow these metadata fields
when writing prompts, reference PL/SQL, tests, and annotations. In particular,
they must ensure that the reference answer matches the user request, compiles and
runs under the provided Oracle-dialect DDL/DML, uses the required constructs,
and maintains turn-to-turn consistency for multi-turn examples. Metadata fields
such as \texttt{tables\_required}, \texttt{columns\_required},
\texttt{reasoning\_types}, and \texttt{plsql\_constructs} are checked for
accuracy and completeness.

\subsection{Spider 2 Normalization to Oracle Dialect}
\label{app:spider2-oracle-normalization}

Spider 2.0 Lite databases were normalized to Oracle DB before PL/SQL task
construction and execution. Since the source databases are adapted from
heterogeneous SQL dialects, including SQLite, Snowflake, and BigQuery, we applied
a normalization pipeline that converts source DDL/DML into Oracle-compatible
schema and data artifacts.

\paragraph{Identifier normalization.}
We normalized table, column, and schema identifiers to satisfy Oracle naming
rules. Identifiers longer than Oracle's 128-character limit were truncated.
Names containing spaces, hyphens, dots, or other special characters were
rewritten using underscores to produce safe, unquoted Oracle identifiers.
Identifiers beginning with digits or underscores were prefixed with valid
alphabetic prefixes, such as \texttt{C\_} for columns and \texttt{S\_} for
schemas. We also prefixed identifiers that conflict with Oracle reserved words
to prevent invalid-identifier and invalid-table-name errors during schema
creation.

\paragraph{Type normalization.}
We converted source-specific data types to Oracle-compatible types. For example,
SQLite- and MySQL-style types such as \texttt{INT}, \texttt{BIGINT},
\texttt{FLOAT}, and \texttt{DATETIME} were mapped to Oracle types such as
\texttt{NUMBER(38)}, \texttt{NUMBER}, \texttt{TIMESTAMP}, \texttt{VARCHAR2},
\texttt{CLOB}, \texttt{BLOB}, \texttt{CHAR(1)}, and \texttt{DATE}.

\paragraph{Literal and value normalization.}
We normalized data literals to avoid Oracle loading and execution errors. Date
and timestamp values appearing in ISO-8601 format, U.S. date format, or bare
date format were converted to valid Oracle date or timestamp literals using
\texttt{DATE}, \texttt{TIMESTAMP}, \texttt{TO\_DATE}, or
\texttt{TO\_TIMESTAMP}. MySQL-style backslash-escaped quotes were rewritten
using Oracle's single-quote escaping convention. MySQL-style \texttt{\texttt NULL} values were converted to Oracle \texttt{NULL}. Oversized string literals
exceeding Oracle's \texttt{VARCHAR2} literal limit were split into chunks and
concatenated using \texttt{TO\_CLOB}. Invalid numeric values such as
\texttt{Infinity}, \texttt{-Inf}, and \texttt{NaN} were replaced with
\texttt{NULL} to prevent numeric conversion errors.

\paragraph{BigQuery-specific normalization.}
For Spider 2.0 Lite databases originating from BigQuery, we additionally handled
values that do not have direct Oracle equivalents. In particular, BigQuery
\texttt{GEOGRAPHY} values exported as coordinate tuples were detected and
replaced with \texttt{NULL}, since these values cannot be loaded directly into
standard Oracle numeric columns without a spatial-type conversion layer.

Overall, this normalization step ensures that Spider 2.0 Lite schemas and data
can be instantiated consistently in Oracle DB and used for executable PL/SQL
code generation, code repair, and interactive development tasks.

\subsection{Task Curation Methodology}
\label{app:task-curation-methodology}

Each task in the benchmark is created from preassigned metadata fields that
control the source, scope, and expected PL/SQL behavior. These fields specify
the source dataset, target database, difficulty level, number of turns, required
PL/SQL constructs, and reasoning types. Example constructs include functions,
procedures, packages, cursors, exception handling, object types, dynamic SQL,
and other procedural database features. Example reasoning types include data
retrieval, validation, debugging, control flow, aggregation, object-oriented
design, exception handling, and state-aware procedural logic.

\paragraph{Metadata-guided prompt construction.}
Annotators write natural-language task prompts that satisfy all assigned
metadata fields. Prompts are required to reflect realistic PL/SQL workflows, such as
reporting, auditing, validation, debugging, business-rule enforcement,
and database-grounded procedural processing. Tasks may require models
to reason about data-processing logic, but their target behavior does not
require persistent modifications to database state. For schema-grounded tasks, prompts
must align with the assigned database and use the relevant tables and columns.
For MBPP-derived tasks, prompts are paired with PL/SQL-compatible inputs,
outputs, and Oracle types when needed.

\paragraph{Task families.}
The benchmark contains three task families. In code generation tasks, the model generates a new PL/SQL function,
procedure, package, or related program unit from a natural-language
specification. In code repair tasks, the model is
given incorrect, incomplete, or failing PL/SQL code and must repair it while
preserving the intended behavior. In interactive development tasks, the model
receives a sequence of user requests over multiple turns. Later turns may add
new requirements, modify business logic, introduce debugging feedback, or ask
for repairs, and the model must preserve context from earlier turns.

\paragraph{Difficulty assignment.}
Task difficulty is assigned according to schema complexity, PL/SQL construct
coverage, reasoning requirements, and interaction structure. Simple tasks use
limited schema context and basic PL/SQL constructs. Intermediate tasks involve
richer schema access, joins, aggregation, loops, cursors or exception handling. Advanced tasks combine multiple tables, complex business logic,
object types, dynamic SQL, package-level structure, exception handling,
or explicit debugging and cross-object requirements. Multi-turn tasks are additionally
scored by whether the interaction requires state tracking, consistent naming,
incremental modification, or repair of earlier code.

\paragraph{Reference solutions and tests.}
Each task is paired with a reference PL/SQL solution and executable validation
tests. Reference solutions must match the prompt exactly, compile in Oracle DB,
and run against the task schema. Tests must be deterministic and runnable
without modification. For multi-turn tasks, the reference solution and tests are
updated consistently across turns so that later requirements do not contradict
or silently break earlier behavior.

\paragraph{Coverage tracking.}
During curation, we track the source dataset, database, required tables,
required columns, reasoning types, PL/SQL constructs, difficulty level, and turn
count. These annotations are used to balance benchmark coverage across schemas,
task types, PL/SQL features, and reasoning categories, and they enable
fine-grained model error analysis after evaluation.

\subsection{Spider 1.0 Adaptation}
\label{app:spider1-plsql-conversion}

For Spider-PLSQL, we adapt the Oracle-converted release of Spider 1.0
\citep{spider_oracle_conversion}, which provides the original natural-language
questions together with corresponding SQL queries in Oracle syntax. We retain
the original questions and use the Oracle SQL result set as the reference for
execution accuracy. We instruct the model to format PL/SQL output so that it
can be compared directly with the textual gold-SQL results, using
\texttt{DBMS\_OUTPUT.PUT\_LINE} with one result row per line and comma-separated
field values. These additional output-format instructions are shown in
Figure~\ref{fig:sample_spider1_prompt}.

\begin{figure*}[!htbp]
\small
\begin{Verbatim}[breaklines=true,breakanywhere=true]

Output requirements:
- Produce result rows only with DBMS_OUTPUT.PUT_LINE; do not use RETURN, OUT parameters, ref cursors, OPEN ... FOR, PIPE ROW, SELECT output, table writes, or any other output channel as the final answer.
- Print no headers, labels, trailers, summaries, or explanatory text.
- Print exactly one output line per result row.
- For multi-column rows, separate fields only with commas; do not use spaces, tabs, pipes, semicolons, or JSON/list formatting as delimiters.
- For single-column rows, print only the value.
- Format DATE values as YYYY-MM-DD.
- Format TIMESTAMP values as YYYY-MM-DD HH24:MI:SS.
- Print string field values exactly as text without surrounding single quotes or double quotes.
\end{Verbatim}
\caption{Output-format instructions for Spider-PLSQL.}
\label{fig:sample_spider1_prompt}
\end{figure*}

\begin{table}[t]
\centering
\scriptsize
\setlength{\tabcolsep}{3pt}
\renewcommand{\arraystretch}{1.03}
\begin{tabularx}{\columnwidth}{@{}p{0.34\columnwidth}X@{}}
\toprule
\textbf{Quality Dimension} & \textbf{Check} \\
\midrule
Prompt--answer alignment &
Required PL/SQL features, such as procedures, cursors, exceptions,
object types, or dynamic SQL. \\
\addlinespace[2.5pt]

Metadata accuracy &
Required tables/columns, difficulty, reasoning types, and PL/SQL features are correct. \\
\addlinespace[2.5pt]

Syntax and compilation &
Reference PL/SQL compiles in the target Oracle environment. \\
\addlinespace[2.5pt]

Execution correctness &
Program output matches the expected result. \\
\addlinespace[2.5pt]

Test-case validity &
Tests run without modification and validate the reference answer. \\
\addlinespace[2.5pt]

Idiomatic PL/SQL &
Code is readable, maintainable, Oracle-dialect compliant, and avoids anti-patterns. \\
\addlinespace[2.5pt]

Feature coverage &
Required PL/SQL features, such as procedures, cursors, exceptions, or dynamic SQL are present. \\
\addlinespace[2.5pt]

Safety &
Tasks avoid unsafe operations and unsupported SQL syntax. \\
\addlinespace[2.5pt]

Multi-turn coherence &
Later turns preserve prior context while adding realistic changes. \\
\addlinespace[2.5pt]

Format validity &
JSON and code fields follow the required format. \\
\addlinespace[2.5pt]

Enterprise relevance &
Prompt reflects realistic PL/SQL workflows such as validation, reporting, debugging, or maintenance. \\
\bottomrule
\end{tabularx}
\caption{Quality dimensions used in \textsc{PLSQLBench} review.}
\label{tab:quality-dimensions}
\end{table}

\begin{figure*}[!htbp]
\small
\centering
\begin{tcolorbox}[colback=white,colframe=black,width=\textwidth,
                  boxrule=0.5pt,arc=0pt,outer arc=0pt,
                  left=6pt,right=6pt,top=6pt,bottom=6pt]

\texttt{def check(candidate):}

\texttt{\hspace{4em}assert candidate([[1, 2, 3], [4, 8, 2], [1, 5, 3]], 2, 2) == 8}

\texttt{\hspace{4em}assert candidate([[2, 3, 4], [5, 9, 3], [2, 6, 4]], 2, 2) == 12}

\texttt{\hspace{4em}assert candidate([[3, 4, 5], [6, 10, 4], [3, 7, 5]], 2, 2) == 16}
\end{tcolorbox}
\caption{A sample MBPP-style test suite.}
\label{fig:mbpp-test-sample}
\end{figure*}

\subsection{MBPP and MBPP+ Python-to-PL/SQL Conversion}
\label{app:mbpp-plsql-conversion}
The Mostly Basic Programming Problems (MBPP) dataset 
has 974 short Python programming assignments designed to be solvable by entry-level programmers and cover programming fundamentals, such as numeric, list, and string manipulations, along with standard library functionality.  \cite{austin2021program}.
Each assignment requests to create a short Python function from a text description.
Correctness of the implementation is verified using unit tests. 

However, \citet{DBLP:conf/nips/LiuXW023} showed that a small number of unit tests can fail to detect a substantial fraction of incorrect implementations. To address this issue, they introduced HumanEval+, an enhanced version of HumanEval \cite{chen2021evaluating} with approximately $80\times$ more unit tests, and observed decreases in pass rates by 9.3--28.9\%. \citet{liu2024evaluating} subsequently introduced MBPP+, a subset of MBPP with a substantially expanded set of unit tests.

We took MBPP and MBPP+ and ``translated'' them into PL/SQL. The resulting
PL/SQL functions do not require persistent database tables; the conversion
focuses on function signatures, inputs, expected outputs, and any Oracle
object or collection types needed to represent structured inputs. We, nevertheless, believe it still provides a useful indication of the model’s PL/SQL coding abilities. 

This conversion was fully rule-based and automatic:  
\begin{itemize}
\item We ingested MBPP-style test case definitions and analyzed them with Python’s AST parser to extract assertions of the form \texttt{assert candidate(<args>) == <expected>} (see Figure ~\ref{fig:mbpp-test-sample} for an example).  

\item The converter then located each \texttt{candidate(...)} call and extracted its literal arguments and expected return value. Using these extracted test cases, it attempted to infer the function's argument types and return type.
The converter rejected test cases containing non-literal or unsafe expressions, as well as tasks for which it could not infer argument types that were consistent across all unit tests.

\item Using the inferred signature, the tool emitted a PL/SQL function scaffold with the exact argument and return types, along with a generated unit-test block that calls the function and asserts equality on each case.  

\item The converter also generated the necessary Oracle collection types (e.g., \texttt{CREATE TYPE} \ldots {AS TABLE OF } \ldots) once per unique element type to ensure the function can accept array-like inputs. 
\item We carried out an Oracle-backed replay validation of the converted benchmark records. For each converted problem, we constructed a replay PL/SQL function that maps the converted test inputs to the corresponding expected outputs and executed the generated PL/SQL unit-test block against this function. This check verifies that the generated PL/SQL test harness is itself executable and that, for every retained record, there exists at least one PL/SQL function that can satisfy the emitted tests.
\end{itemize}
In the final validation pass, 806 out of 823 candidate MBPP records and 308 out of 323 candidate MBPP+ records passed these checks.

While MXEval \cite{DBLP:conf/iclr/AthiwaratkunGWL23} explored a similar high-level approach to constructing multilingual code-generation benchmarks, our work differs in several important respects. In particular, we target PL/SQL language, which is not supported by MXEval.
Furthermore, we introduce a different translation pipeline and evaluation methodology. 
MXEval relies on a rule-based conversion framework for translating prompts and test cases, supplemented by manual expert review of selected languages to identify issues and iteratively improve the conversion procedure. 
In contrast, our conversion pipeline was applied as a fixed rule-based procedure, without iterative manual refinement of the translated benchmark. We then relied on automated replay-based validation, discarding cases for which a valid PL/SQL replay function could not be constructed or could not pass the translated test harness.

\begin{figure}[!htbp]
\small
\centering
\begin{lstlisting}
CREATE OR REPLACE FUNCTION count_first_elements(
  p1 IN CLOB_NTT
) RETURN NUMBER
AS
BEGIN
  IF p1 IS NULL OR p1.COUNT = 0 THEN
    RETURN 0;
  END IF;

  RETURN p1.COUNT - 1;
END;
\end{lstlisting}
\caption{A sample MBPP+ shortcut function generated by a model with Python unit-tests in the prompt.}
\label{fig:mbpp-unittest-shortcut-sample}
\end{figure}

To further corroborate the correctness of the Python-to-PL/SQL translation and evaluation harness, we carried out two tests. In the first test, we generated PL/SQL solutions from prompts containing the Python canonical solution and all available unit-test examples (using GPT-5.5). This experiment was intended to establish a lower bound on the number of converted problems for which the model can generate a valid PL/SQL implementation that passes the translated tests when given sufficient information about the intended computation. We did it only for MBPP+,
because on MBPP the best models already achieve nearly-perfect accuracy.

Because the Python unit tests were visible, a model could potentially pass them using a shortcut rather than implementing the intended function. Figure~\ref{fig:mbpp-unittest-shortcut-sample} shows such an MBPP+ example. 
Thus, we carried out two checks to ensure that shortcut or replay solutions are rare. 
\begin{itemize}
    \item First, we compared generated-function lengths across prompt conditions. The hinted and no-hint solutions were similar in length for MBPP+, with no evidence that exposing reference solutions or unit tests caused generations to collapse into short lookup-style implementations.
    \item Second, we manually reviewed paired generations produced with no hints and with the maximum available hints. We inspected 50 sampled MBPP+ records. 
     Apart from one identified case in which the model appeared to exploit a shortcut, the inspected hinted generations appeared to be bona-fide implementations of the requested functions rather than replay tables or test-specific shortcuts.
\end{itemize}

\begin{table}[!htbp]
\centering
\small
\begin{tabular}{llc}
\toprule
\textbf{Dataset} & \textbf{Prompt condition} & \textbf{Pass rate} \\
\midrule
MBPP+ & No hints                         & 45.1\% \\
MBPP+ & Canonical solution               & 56.8\% \\
MBPP+ & Canonical solution + 20\% tests  & 68.8\% \\
MBPP+ & Canonical solution + 100\% tests & 87.3\% \\
\bottomrule
\end{tabular}
\caption{PL/SQL pass rates for MBPP+ under different prompt conditions.}
\label{tab:mbpp-hinted-pass-rates}
\end{table}

The corresponding pass rates provide the diagnostic lower-bound signal that motivated these experiments.
According to Table~\ref{tab:mbpp-hinted-pass-rates}, providing progressively more information about the intended Python computation substantially increased the fraction of translated problems for which the model could generate a passing PL/SQL implementation.

As a second benchmark-validation check, we compared generated PL/SQL functions with the original Python canonical solutions. For each supported record, we evaluated the generated Oracle function on the converted inputs and the Python reference solution on the corresponding Python inputs, normalized known representation differences, and compared the outputs. To maximize the number of informative comparisons, the PL/SQL functions were generated using the hinted prompts described above.

The PL/SQL outputs matched the Python-reference checks for 96.8\% of output-comparable MBPP records and 89.7\% of MBPP+ records. At the pass/fail level, agreement was 96.7\% for MBPP and 91.6\% for MBPP+, increasing to 98.7\% and 96.6\%, respectively, when only conclusive comparisons were considered (compatibility of a small fraction of outputs could not be verified). Note that perfect agreement is not expected because Python-to-PL/SQL comparison is itself heuristic, especially for complex or nested types whose representations do not map exactly between the two languages. We therefore use this comparison as a strong sanity check rather than as a proof of exact semantic equivalence.

Taken together, the replay validation, Python--PL/SQL behavioral cross-checks, and hinted-generation analysis collectively provide evidence that the retained MBPP and MBPP+ tasks form a reliable benchmark surface for evaluating PL/SQL coding ability with limited database grounding.

\subsection{Spider 2.0 Quality Control Pipeline}
\label{app:review-pipeline}

Quality control combines automatic checks, human review, post-processing, and
LLM-as-judge review. Automatic review is applied before submission and checks all
examples for formatting, metadata completeness, prompt-reference alignment,
schema references, compilation, execution, and test validity. Human reviewers
then inspect examples for semantic correctness, idiomatic PL/SQL, realistic
developer intent, and recurring annotation issues. A post-processing pipeline
checks each delivery file for SQL compilation, test execution, reasoning-type
labels, construct labels, and under- or over-claimed table and column names.
Finally, an LLM-as-judge pass evaluates dimensions such as specification
alignment, control-flow correctness, data handling, edge cases, and PL/SQL
quality.

\subsection{Quality Dimensions}
\label{app:quality-dimensions}

Table~\ref{tab:quality-dimensions} summarizes the primary quality dimensions
used during review.

% metric definition
\section{Metric Definitions}
\label{app:metrics}

All metrics are computed from execution-based unit tests using a single
generated PL/SQL program per task or turn (\emph{Pass@1}). A compilation error,
runtime error, or failure to create the required database object is treated as
passing zero tests for the affected task or turn. Since tasks may contain
different numbers of unit tests, we compute test-level partial credit within
each task or episode first, then average across tasks or episodes.

\paragraph{Single-turn metrics.}
For a single-turn task $i$, let $n_i$ be the number of unit tests and let $p_i$
be the number of tests passed by the generated program. We define the task-level
test pass rate as:

\[
\mathrm{TestPass}_i = \frac{p_i}{n_i}.
\]

The strict suite-level score is:

\[
\mathrm{SuitePass}_i =
\begin{cases}
1 & \text{if } p_i = n_i,\\
0 & \text{otherwise.}
\end{cases}
\]

For a set of single-turn tasks $\mathcal{S}$, we report:

\[
\mathrm{MeanTestPass@1}
=
\frac{1}{|\mathcal{S}|}
\sum_{i \in \mathcal{S}} \mathrm{TestPass}_i,
\]

\[
\mathrm{SuitePass@1}
=
\frac{1}{|\mathcal{S}|}
\sum_{i \in \mathcal{S}} \mathrm{SuitePass}_i.
\]

Thus, Mean Test Pass@1 gives each task equal weight after normalizing by its
number of unit tests. For example, if one task passes 2 of 3 tests and another
passes 1 of 3 tests, then
$\mathrm{MeanTestPass@1} = ((2/3) + (1/3))/2 = 1/2$.

\paragraph{Multi-turn metrics.}
For a multi-turn episode $e$, let $m_e$ be the number of turns. At turn $t$, let
$n_{e,t}$ be the number of unit tests and let $p_{e,t}$ be the number of tests
passed. We first define whether each turn is fully solved:

\[
\mathrm{TurnSolved}_{e,t} =
\begin{cases}
1 & \text{if } p_{e,t} = n_{e,t},\\
0 & \text{otherwise.}
\end{cases}
\]

Episode Pass requires every turn to be fully solved:

\[
\mathrm{EpisodePass}_e =
\begin{cases}
1 & \text{if } \sum_{t=1}^{m_e} \mathrm{TurnSolved}_{e,t} = m_e,\\
0 & \text{otherwise.}
\end{cases}
\]

Turn Suite Pass measures the fraction of turns that are fully solved:

\[
\mathrm{TurnSuitePass}_e =
\frac{1}{m_e}
\sum_{t=1}^{m_e} \mathrm{TurnSolved}_{e,t}.
\]

Mean Test Pass gives partial credit across all tests in an episode:

\[
\mathrm{EpisodeTestPass}_e =
\frac{
\sum_{t=1}^{m_e} p_{e,t}
}{
\sum_{t=1}^{m_e} n_{e,t}
}.
\]

For a set of multi-turn episodes $\mathcal{E}$, we report:

\[
\mathrm{MeanTestPass@1}
=
\frac{1}{|\mathcal{E}|}
\sum_{e \in \mathcal{E}} \mathrm{EpisodeTestPass}_e,
\]

\[
\mathrm{EpisodePass@1}
=
\frac{1}{|\mathcal{E}|}
\sum_{e \in \mathcal{E}} \mathrm{EpisodePass}_e,
\]

\[
\mathrm{TurnSuitePass@1}
=
\frac{1}{|\mathcal{E}|}
\sum_{e \in \mathcal{E}} \mathrm{TurnSuitePass}_e.
\]

Thus, each episode contributes equally to the final multi-turn benchmark score,
even when episodes contain different numbers of unit tests. For example, for one row of data, in a
three-turn episode with six total tests, if the model fully solves two turns and
passes 5 of 6 tests overall, then $\mathrm{EpisodePass}=0$,
$\mathrm{TurnSuitePass}=2/3$, and $\mathrm{EpisodeTestPass}=5/6$.

%%%%%%%%%%%%%%%%%%%%%%%%%%%%%%%%%%%%%%%%%%%%%%%%%%%%%%%%%%%%%%%%%%%%%%%%%%%%%%
% Extra results
%%%%%%%%%%%%%%%%%%%%%%%%%%%%%%%%%%%%%%%%%%%%%%%%%%%%%%%%%%%%%%%%%%%%%%%%%%%%%%

\clearpage
\onecolumn
\clearpage
\onecolumn

\section{Dataset Statistics Details}
\label{app:data_stats_details}

\subsection{Turn Distribution of Spider2-MT}
Spider2-MT contains 271 complete multi-turn conversations spanning 978 turns
across the development and test partitions. Each conversation contains three
to five turns. Three-turn conversations form the largest group, with 155
conversations (57.2\%), followed by 67 four-turn conversations (24.7\%) and
49 five-turn conversations (18.1\%). The development partition contains 208
conversations and 750 turns, while the test partition contains 63 conversations
and 228 turns.
\begin{table}[H]
\centering
\small
\setlength{\tabcolsep}{4pt}
\renewcommand{\arraystretch}{1.05}
\begin{tabular}{lrrrrrr}
\toprule
\textbf{Conversation Length} & \textbf{Dev Conv.} & \textbf{Dev Turns} & \textbf{Test Conv.} & \textbf{Test Turns} & \textbf{All Conv.} & \textbf{All Turns} \\
\midrule
3-turn & 117 & 351 & 38 & 114 & 155 & 465 \\
4-turn & 56 & 224 & 11 & 44 & 67 & 268 \\
5-turn & 35 & 175 & 14 & 70 & 49 & 245 \\
\midrule
\textbf{Total} & \textbf{208} & \textbf{750} & \textbf{63} & \textbf{228} & \textbf{271} & \textbf{978} \\
\bottomrule
\end{tabular}
\caption{Turn-length distribution of Spider2-MT conversations in
\textsc{PLSQLBench}.}
\label{tab:turn-distribution}
\end{table}

\subsection{Reasoning Type and Difficulty Distribution on Spider2}
\label{app:spider2_reasoning_difficulty}

We further characterize the Spider2-ST and Spider2-MT splits by difficulty and
reasoning type. Difficulty labels are assigned according to the procedural and
database reasoning required by each task. \textit{Simple} tasks typically involve a
single retrieval or straightforward control-flow pattern over a small number of
tables, with limited state or exception behavior. \textit{Intermediate} tasks
combine multiple procedural requirements, such as joins, aggregation, cursor
iteration, validation logic, or structured exception handling. \textit{Advanced}
tasks require more complex PL/SQL program construction, such as packages,
multi-step stateful logic, dynamic SQL, object-oriented types, bulk processing, or
coordinated behavior across turns in a multi-turn conversation.

\begin{table}[H]
\centering
\small
\setlength{\tabcolsep}{4pt}
\renewcommand{\arraystretch}{1.05}
\begin{tabular}{lrrrr}
\toprule
\textbf{Difficulty} & \textbf{ST Dev} & \textbf{ST Test} & \textbf{MT Dev} & \textbf{MT Test} \\
\midrule
Simple       & 229 (56.3) & 65 (63.1) & 459 (61.2) & 134 (58.8) \\
Intermediate & 169 (41.5) & 37 (35.9) & 254 (33.9) & 66 (28.9) \\
Advanced     & 9 (2.2) & 1 (1.0) & 37 (4.9) & 28 (12.3) \\
\bottomrule
\end{tabular}
\caption{Difficulty distribution for Spider2-ST and Spider2-MT dev/test
splits. Parenthesized values are percentages within each split; Spider2-MT
counts are over turns.}
\label{tab:spider2-difficulty-split}
\end{table}

\begin{table}[H]
\centering
\scriptsize
\setlength{\tabcolsep}{4pt}
\renewcommand{\arraystretch}{1.05}
\begin{tabular}{lrrrr}
\toprule
\textbf{Reasoning Type} &
\textbf{Spider2-ST Dev} &
\textbf{Spider2-ST Test} &
\textbf{Spider2-MT Dev} &
\textbf{Spider2-MT Test} \\
\midrule
Exception Handling & 407 (100.0) & 103 (100.0) & 750 (100.0) & 226 (99.1) \\
Data Retrieval & 407 (100.0) & 103 (100.0) & 746 (99.5) & 227 (99.6) \\
Control Flow & 398 (97.8) & 102 (99.0) & 716 (95.5) & 221 (96.9) \\
Debugging & 336 (82.6) & 81 (78.6) & 665 (88.7) & 210 (92.1) \\
Iterative Processing & 283 (69.5) & 66 (64.1) & 436 (58.1) & 143 (62.7) \\
Aggregation & 162 (39.8) & 38 (36.9) & 330 (44.0) & 116 (50.9) \\
Structural / Type Reasoning & 111 (27.3) & 27 (26.2) & 268 (35.7) & 89 (39.0) \\
Validation & 108 (26.5) & 26 (25.2) & 183 (24.4) & 57 (25.0) \\
Cursors & 101 (24.8) & 20 (19.4) & 142 (18.9) & 67 (29.4) \\
Decision Logic & 60 (14.7) & 10 (9.7) & 107 (14.3) & 40 (17.5) \\
\bottomrule
\end{tabular}
\caption{Top-10 reasoning-type distribution for Spider2-ST and Spider2-MT
dev/test splits, ranked by Spider2-MT dev frequency. Counts are multi-label;
parenthesized values are percentages within each split, and Spider2-MT counts
are over turns.}
\label{tab:spider2-reasoning-top10}
\end{table}

\onecolumn

\lstdefinestyle{plsqlbench}{
  basicstyle=\ttfamily\footnotesize,
  breaklines=true,
  columns=fullflexible,
  keepspaces=true,
  showstringspaces=false,
  frame=none
}

\newtcolorbox{benchmarkexample}[1]{
  enhanced,
  breakable,
  colback=gray!3,
  colframe=gray!35,
  boxrule=0.4pt,
  arc=2pt,
  left=6pt,
  right=6pt,
  top=6pt,
  bottom=6pt,
  title=#1,
  fonttitle=\bfseries,
  coltitle=black,
  colbacktitle=gray!15
}

\subsection{Representative Benchmark Instances}
\label{app:data-example}

We provide one representative instance from each major \textsc{PLSQLBench}
subset. These examples illustrate the range of artifacts required by the
benchmark: schema-grounded single-turn program units, MBPP-derived function generation
without persistent database tables, and multi-turn procedure revision.

\paragraph{Spider2-ST.}
The Spider2-ST subset contains enterprise-style schema-grounded PL/SQL tasks.
The following instance asks for a reusable function with anchored database
behavior and explicit exception handling.

\begin{benchmarkexample}{Spider2-ST single-turn example}
\textbf{Instance:} \texttt{spider2\_single\_turn\_2} \\
\textbf{Database:} \texttt{city\_legislation}

\medskip
\textbf{Task.}
Design and implement a reusable PL/SQL function named
\texttt{sf\_get\_city\_population} that retrieves the population information
for a given city.

\medskip
\textbf{Requirements.}
\begin{itemize}[leftmargin=1.5em, itemsep=1pt, topsep=2pt]
  \item Function name: \texttt{sf\_get\_city\_population}
  \item Parameter: \texttt{p\_city\_name IN VARCHAR2}
  \item Return value: \texttt{VARCHAR2} population text
  \item Match city names case-insensitively.
  \item Return \texttt{City name cannot be NULL} for NULL input.
  \item Return \texttt{City not found} for \texttt{NO\_DATA\_FOUND}.
  \item Return \texttt{Duplicate city records found} for \texttt{TOO\_MANY\_ROWS}.
  \item Return \texttt{Unexpected error occurred} for other failures.
\end{itemize}

\medskip
\textbf{Example tests.}
\begin{lstlisting}[style=plsqlbench]
SELECT sf_get_city_population('Atlanta') FROM DUAL;
SELECT sf_get_city_population(NULL) FROM DUAL;
\end{lstlisting}
\end{benchmarkexample}

\paragraph{MBPP-PLSQL.}
The MBPP-PLSQL subset converts MBPP programming tasks into PL/SQL
function-generation tasks. These tasks do not require persistent database
tables; instead, they provide the necessary PL/SQL types and unit tests.

\begin{benchmarkexample}{MBPP-PLSQL function generation example}
\textbf{Instance:} \texttt{MBPP/1}

\medskip
\textbf{Task.}
Complete the Oracle PL/SQL function \texttt{min\_cost}.

\medskip
\textbf{Original Python specification.}
Write a function to find the minimum cost path to reach \texttt{(m, n)} from
\texttt{(0, 0)} for the given cost matrix \texttt{cost[][]}.

\medskip
\textbf{Examples.}
\begin{lstlisting}[style=plsqlbench]
min_cost([[1, 2, 3], [4, 8, 2], [1, 5, 3]], 2, 2) -> 8
min_cost([[2, 3, 4], [5, 9, 3], [2, 6, 4]], 2, 2) -> 12
min_cost([[3, 4, 5], [6, 10, 4], [3, 7, 5]], 2, 2) -> 16
\end{lstlisting}

\medskip
\textbf{Provided PL/SQL collection types.}
\begin{lstlisting}[style=plsqlbench]
CREATE TYPE NUMBER_NTT AS TABLE OF NUMBER;
CREATE TYPE NUMBER_NTT_NTT AS TABLE OF NUMBER_NTT;
\end{lstlisting}

\medskip
\textbf{Function signature.}
\begin{lstlisting}[style=plsqlbench]
CREATE OR REPLACE FUNCTION min_cost(
  p1 IN NUMBER_NTT_NTT,
  p2 IN NUMBER,
  p3 IN NUMBER
) RETURN NUMBER
\end{lstlisting}
\end{benchmarkexample}

\paragraph{Spider2-MT.}
The Spider2-MT subset evaluates whether models can revise and extend prior
PL/SQL artifacts across turns. The following conversation shows a three-turn
revision sequence over the same stored procedure.

\begin{benchmarkexample}{Spider2-MT multi-turn example}
\textbf{Conversation:} \texttt{plsql-multiturn-0002} \\
\textbf{Database:} \texttt{stacking} \\
\textbf{Complexity:} advanced \\
\textbf{Source dataset:} Spider 2.0 Lite \\
\textbf{Turns:} 3

\medskip
\textbf{Shared procedure interface.}
\begin{lstlisting}[style=plsqlbench]
SP_GET_MODEL_PERFORMANCE(
  p_model    IN VARCHAR2,
  p_analysis IN VARCHAR2
)
\end{lstlisting}

\medskip
\textbf{Turn 1: create the procedure.}

\textit{User request.}
Create \texttt{SP\_GET\_MODEL\_PERFORMANCE} to retrieve and print the latest
test score for a model and analysis from \texttt{STACKING.MODEL\_SCORE}.

\medskip
\textit{Relevant columns.}
\begin{lstlisting}[style=plsqlbench]
STACKING.MODEL_SCORE:
  MODEL, VERSION, NAME, TEST_SCORE, STEP
\end{lstlisting}

\textit{Required behavior.}
\begin{itemize}[leftmargin=1.5em, itemsep=1pt, topsep=2pt]
  \item Match \texttt{MODEL} and \texttt{NAME} case-insensitively with \texttt{LOWER(...)}.
  \item Select the latest row using \texttt{ORDER BY VERSION DESC, STEP DESC} and \texttt{ROWNUM = 1}.
  \item Print \texttt{Model: <p\_model> | Score: <lv\_score>}.
  \item Handle \texttt{NO\_DATA\_FOUND} with a \texttt{DBMS\_OUTPUT} message.
  \item Handle unexpected failures with \texttt{Unexpected error occurred}.
\end{itemize}

\textit{Representative tests.}
\begin{lstlisting}[style=plsqlbench]
EXEC SP_GET_MODEL_PERFORMANCE('RFCE', 'Diabetes');
-- Expected: Model: RFCE | Score: 1

EXEC SP_GET_MODEL_PERFORMANCE('Unknown Model', 'analysis_999');
-- Expected: Record not found for Model: Unknown Model in Analysis: analysis_999

EXEC SP_GET_MODEL_PERFORMANCE('RFCE', 'Pain');
-- Expected: Record not found for Model: RFCE in Analysis: Pain

EXEC SP_GET_MODEL_PERFORMANCE(NULL, NULL);
-- Expected: Record not found for Model:  in Analysis:
\end{lstlisting}

\medskip
\textbf{Turn 2: modify the decision logic.}

\textit{Conversation dependency.}
The model must preserve the procedure name, parameters, score lookup, version
ordering, and exception behavior from Turn 1, then replace the output logic with
threshold-based branching.

\medskip
\textit{User request.}
Update \texttt{SP\_GET\_MODEL\_PERFORMANCE} to evaluate the retrieved score
against a \texttt{0.7} threshold and print distinct success or warning messages.

\medskip
\textit{Added behavior.}
\begin{lstlisting}[style=plsqlbench]
If lv_score >= 0.7:
  Success: Model <p_model> passed with score <lv_score>

If lv_score < 0.7:
  Warning: Model <p_model> failed to meet the 0.7 threshold.
\end{lstlisting}

\textit{Representative tests.}
\begin{lstlisting}[style=plsqlbench]
EXEC SP_GET_MODEL_PERFORMANCE('Stack', 'Pumpkin Seeds');
-- Expected: Success: Model Stack passed with score 1

EXEC SP_GET_MODEL_PERFORMANCE('MLPC1', 'iris');
-- Expected: Warning: Model MLPC1 failed to meet the 0.7 threshold.

EXEC SP_GET_MODEL_PERFORMANCE('Unknown Model', 'analysis_999');
-- Expected: Record not found for Model: Unknown Model in Analysis: analysis_999

EXEC SP_GET_MODEL_PERFORMANCE(NULL, NULL);
-- Expected: Record not found for Model:  in Analysis:
\end{lstlisting}

\medskip
\textbf{Turn 3: add a join and object-oriented formatting.}

\textit{Conversation dependency.}
The model must preserve the Turn 2 threshold behavior while extending the
procedure with a join to \texttt{STACKING.PROBLEM} and introducing
object-oriented formatting through a SQL object type and type body.

\medskip
\textit{User request.}
Update \texttt{SP\_GET\_MODEL\_PERFORMANCE} to join
\texttt{STACKING.MODEL\_SCORE} and \texttt{STACKING.PROBLEM} on \texttt{NAME},
fetch the problem \texttt{TYPE}, include \texttt{lv\_type} in all outputs, and
implement \texttt{MODEL\_PERFORMANCE\_OT} plus its type body.

\medskip
\textit{New artifacts.}
\begin{itemize}[leftmargin=1.5em, itemsep=1pt, topsep=2pt]
  \item \texttt{CREATE OR REPLACE TYPE MODEL\_PERFORMANCE\_OT AS OBJECT (...)}
  \item \texttt{CREATE OR REPLACE TYPE BODY MODEL\_PERFORMANCE\_OT}
  \item \texttt{CREATE OR REPLACE PROCEDURE SP\_GET\_MODEL\_PERFORMANCE}
\end{itemize}

\textit{Additional schema elements.}
\begin{lstlisting}[style=plsqlbench]
STACKING.PROBLEM:
  NAME, TYPE
\end{lstlisting}

\textit{Added PL/SQL constructs.}
\begin{lstlisting}[style=plsqlbench]
CREATE OR REPLACE TYPE
CREATE OR REPLACE TYPE BODY
MEMBER FUNCTION
INNER JOIN
NVL
TO_CHAR
TOO_MANY_ROWS exception handling
\end{lstlisting}

\textit{Representative tests.}
\begin{lstlisting}[style=plsqlbench]
EXEC SP_GET_MODEL_PERFORMANCE('ABC', 'water quality');
-- Expected: Model: ABC | Analysis: water quality | Type: classification | Score: 1

EXEC SP_GET_MODEL_PERFORMANCE('GPRQ', 'concrete');
-- Expected: Threshold failed for regression model: GPRQ

EXEC SP_GET_MODEL_PERFORMANCE('XGBoost', 'non_existent_analysis');
-- Expected: Record not found for Model: XGBoost in Analysis: non_existent_analysis

EXEC SP_GET_MODEL_PERFORMANCE(NULL, NULL);
-- Expected: Record not found for Model:  in Analysis:
\end{lstlisting}

\medskip
\textbf{Why this is multi-turn.}
Each later turn relies on the earlier procedure interface and lookup semantics,
then asks the model to modify selected parts of the program while preserving
previously established behavior.
\end{benchmarkexample}

\clearpage
\section{Additional Evaluation Results}
\label{app:additional_results}

The main paper reports Mean Test Pass@1 as the primary metric. This appendix
provides additional strict evaluation metrics, including Single-Turn Suite
Pass@1, Multi-Turn Episode Pass@1, and Multi-Turn Turn Suite Pass@1.

%%%%%%%%%%%%%%%%%%%%%%%%%%%%%%%%%%%%%%%%%%%%%%%%%%%%%%%%%%%%%%%%%%%%%%%%%%%%%%
\begin{table}[h]
\centering
\small
\setlength{\tabcolsep}{2.5pt}
\renewcommand{\arraystretch}{1.08}
\begin{tabular}{lccc|cc|c}
\toprule
& \multicolumn{3}{c|}{\textbf{Development}}
& \multicolumn{2}{c|}{\textbf{Test}}
& \textbf{Overall} \\
\cmidrule(lr){2-4}
\cmidrule(lr){5-6}
\cmidrule(l){7-7}
\textbf{Model}
& MBPP
& Spider-PLSQL
& Spider2-ST
& MBPP+
& Spider2-ST
& \shortstack{Test\\Mean} \\
\midrule

\rowcolor{gray!15}
\multicolumn{7}{l}{\textit{Open-weight models}} \\
Llama-4-Maverick & 57.82 & 75.46 & 42.75 & 26.30 & 50.49 & 38.39 \\
Gemma-4-31B & 80.40 & 81.44 & 55.77 & 39.94 & 61.17 & 50.55 \\

\rowcolor{gray!15}
\multicolumn{7}{l}{\textit{Proprietary models}} \\
Gemini-2.5-Flash-Lite & 63.77 & 73.92 & 45.45 & 33.44 & 47.57 & 40.51 \\
Grok-4.3 & 87.59 & 64.95 & 52.83 & 38.31 & 53.40 & 45.85 \\
GPT-5.4-Mini & 70.35 & 76.60 & 46.19 & 38.96 & 62.14 & 50.55 \\
GPT-5.4 & 83.75 & 78.45 & 57.99 & 42.21 & 68.93 & 55.57 \\
GPT-5.6-Sol & \textbf{96.28} & 79.59 & 45.70 & 46.75 & 65.05 & 55.90 \\
Claude-Opus-4.8 & 90.32 & \textbf{84.33} & 56.02 & \textbf{47.73} & 60.19 & 53.96 \\

\rowcolor{gray!15}
\multicolumn{7}{l}{\textit{Tool-augmented agents}} \\
GPT-5.4 Mini + Codex Agent & -- & -- & 56.51 & 45.13 & 64.08 & 54.60 \\
GPT-5.6-Sol + Codex Agent & -- & -- & \textbf{60.93} & 46.10 & \textbf{74.76} & \textbf{60.43} \\
\bottomrule
\end{tabular}
\caption{
Single-turn Suite Pass@1 (\%), with overall means computed only across
the two test sets. Codex Agent rows use medium
reasoning; unavailable evaluations are shown as --. Best results in each
column are shown in bold.
}
\label{tab:single-turn-suite}
\end{table}
%%%%%%%%%%%%%%%%%%%%%%%%%%%%%%%%%%%%%%%%%%%%%%%%%%%%%%%%%%%%%%%%%%%%%%%%%%%%%%
\subsection{Multi-Turn Strict Metrics}
\label{app:mt_strict_metrics}

Table~\ref{tab:mt-strict-metrics} reports Episode Pass@1 and Turn Suite Pass@1
for Spider2-MT. Episode Pass@1 requires every turn in a conversation to pass
all required unit tests, while Turn Suite Pass@1 measures suite-level success
at the individual turn level.

\begin{table}[h]
\centering
\small
\setlength{\tabcolsep}{4pt}
\renewcommand{\arraystretch}{1.08}
\begin{tabular}{lcc|cc}
\toprule
& \multicolumn{2}{c|}{\textbf{Episode Pass@1}}
& \multicolumn{2}{c}{\textbf{Turn Suite Pass@1}} \\
\cmidrule(lr){2-3}
\cmidrule(lr){4-5}
\textbf{Model}
& \textbf{Dev}
& \textbf{Test}
& \textbf{Dev}
& \textbf{Test} \\
\midrule

\rowcolor{gray!15}
\multicolumn{5}{l}{\textit{Open-weight models}} \\
Llama-4-Maverick & 17.31 & 15.87 & 36.93 & 39.47 \\
Gemma-4-31B & 26.44 & 30.16 & 50.53 & 54.39 \\

\rowcolor{gray!15}
\multicolumn{5}{l}{\textit{Proprietary models}} \\
Gemini-2.5-Flash-Lite & 15.38 & 19.05 & 35.07 & 42.98 \\
Grok-4.3 & 23.56 & 28.57 & 48.13 & 51.75 \\
GPT-5.4-Mini & 24.04 & 23.81 & 46.93 & 50.88 \\
GPT-5.4 & 27.88 & 33.33 & 52.53 & 59.21 \\
GPT-5.6-Sol & 23.56 & 22.22 & 47.20 & 50.00 \\
Claude-Opus-4.8 & 26.92 & 28.57 & 50.80 & 52.63 \\

\rowcolor{gray!15}
\multicolumn{5}{l}{\textit{Tool-augmented agents}} \\
GPT-5.4 Mini + Codex Agent & 28.37 & 28.57 & 51.33 & 56.14 \\
GPT-5.6-Sol + Codex Agent & \textbf{33.17} & \textbf{41.27} & \textbf{55.87} & \textbf{64.91} \\
\bottomrule
\end{tabular}
\caption{
Spider2-MT strict multi-turn metrics (\%). Episode Pass@1 requires every turn
in a conversation to pass all unit tests, while Turn Suite Pass@1 measures
suite-level success at the individual turn level. Codex Agent rows use medium reasoning. Best results are
shown in bold.
}
\label{tab:mt-strict-metrics}
\end{table}

% error analysis
% \clearpage

% Preamble

\tcbset{
  errorcase/.style={
    enhanced,
    breakable,
    colback=white,
    colframe=black,
    boxrule=0.5pt,
    arc=0pt,
    outer arc=0pt,
    left=6pt,
    right=6pt,
    top=5pt,
    bottom=5pt,
    before skip=4pt,
    after skip=8pt,
    fontupper=\small,
    before upper={
      \setlength{\parskip}{2pt}
      \setlength{\parindent}{0pt}
    }
  }
}

% Compact centered monospace line. Handles underscores safely.
\newcommand{\errline}[1]{%
  \par\smallskip
  \noindent
  {\centering
    \begin{minipage}{0.96\linewidth}
    \centering\footnotesize\ttfamily
    \detokenize{#1}
    \end{minipage}
    \par
  }%
  \smallskip
}

%%%%%%%%%%%%%%%%%%%%%%%%%%%%%%%%%%%%%%%%%%%%%%%%%%%%%%%%%%%%%%%%%%%%%%%%%%%%%%
\clearpage
\onecolumn
\subsection{Error Analysis Protocol}
\label{app:error_analysis_protocol}

We perform error analysis over the execution-level evaluation logs produced by
the PL/SQL harness. Each evaluated single-turn instance, and each turn in a
multi-turn conversation, is treated as one analysis unit. We first exclude
\texttt{reference\_error} rows, which correspond to reference-solution or harness
execution failures rather than model-attributable failures. For the remaining
rows, we separate passed executions from model failures and assign each failure
to a coarse category using the row status, Oracle diagnostics, skipped-execution
reasons, and expected-versus-actual outputs.

Our taxonomy contains three major model-attributable failure categories. First,
\emph{wrong procedure logic or output} covers generated PL/SQL that compiles or
executes but produces incorrect observable behavior, including incorrect
procedural control flow, exception handling, state updates, output ordering,
formatting required by the task, or query results. Second, \emph{invalid or
incomplete PL/SQL artifacts} covers malformed, incomplete, or non-executable
program units, including syntax errors, truncated procedures, invalid compiled
objects, explanatory text emitted instead of code, and generated SQL skipped by
the safety checker. Third, \emph{interface, signature, or schema grounding
errors} covers failures where the generated artifact does not match the expected
database-facing contract, such as missing entry points, wrong parameter
signatures, missing package members, invalid object names, or incorrect
table/column references.

We use Oracle diagnostics as evidence for assigning failures to these
categories. For example, \texttt{PLS-00905} indicates that a generated program
unit compiled into an invalid object, \texttt{PLS-00201} often indicates a
missing expected entry point, \texttt{PLS-00306} indicates a wrong number or type
of arguments, and \texttt{ORA-00904} or \texttt{ORA-00942} indicate schema or
identifier grounding failures. Output mismatches without Oracle exceptions are
inspected through expected-versus-actual outputs.

For multi-turn data, failures are analyzed at the turn level while retaining the
conversation identifier. This lets us identify failures that arise after
accumulated conversational context. We report aggregate counts by error category,
dataset, and model. We then manually inspect representative examples for each
major category. During manual inspection, we avoid examples that are primarily
artifacts of brittle exact-string evaluation, such as harmless date-format
differences, and instead select cases where the generated PL/SQL exhibits a
clear procedural, artifact-construction, or interface-grounding failure.

\subsection{Representative Error Examples}
\label{app:error_examples}

Figures~\ref{fig:error-logic-output}--\ref{fig:error-interface-schema}
provide representative examples from the three error categories discussed in
Section~\ref{sec:error_analysis}. We include compact execution evidence rather
than full traces to show how each failure is identified from the generated
artifact and unit-test outcome.

\noindent
\begin{tcolorbox}[errorcase]
\textbf{Error Category: Wrong procedure logic or output.}

\textbf{Example.}
Dataset: \texttt{Spider2-MT};
Model: \texttt{GPT-5.4};
Instance: \texttt{plsql-multiturn-0053\_q01};
Tests passed: \texttt{0/3}.

\textbf{Task.}
The user asks for a reusable reporting procedure,
\texttt{sp\_report\_team\_appearances}, that summarizes a team's lineup usage
for a given season. The procedure takes \texttt{p\_team\_id} and
\texttt{p\_year} as inputs, prints up to 15 players ordered by total games, and
handles \texttt{NO\_DATA\_FOUND} by printing:

\errline{No appearances found for given team and year}

\textbf{Generated behavior.}
The generated procedure compiles and contains a cursor loop over aggregated
appearance records. However, it only prints player rows inside the loop. If the
cursor returns no rows, it raises \texttt{NO\_DATA\_FOUND} after the loop and
jumps directly to the exception handler:

\errline{IF v_row_count = 0 THEN RAISE NO_DATA_FOUND; END IF;}

As a result, the generated procedure prints only the exception message for
empty-result cases.

\textbf{Reference behavior.}
The reference procedure treats the routine as a report. It prints report header
lines before iterating through the cursor:

\errline{TEAM=<team_id> YEAR=<year>}
\errline{PLAYER_ID       TOTAL_G   TOTAL_GS}

Only after producing these report headers does it raise
\texttt{NO\_DATA\_FOUND} when no player rows are found. Thus, even in an
empty-result case, the expected observable output includes both the report
context and the exception message.

\textbf{Failure signal.}
For the test case \texttt{p\_team\_id = NYA} and \texttt{p\_year = 2000}, the
expected output is:

\errline{TEAM=NYA YEAR=2000}
\errline{PLAYER_ID       TOTAL_G   TOTAL_GS}
\errline{No appearances found for given team and year}

The generated procedure instead outputs only:

\errline{No appearances found for given team and year}

The same pattern occurs for all three tests. No compile-time or runtime
exception is recorded; the failure is an output mismatch.

\textbf{Why this is an error.}
This is a PL/SQL-specific control-flow and side-effect failure. The model does
not simply choose the wrong SQL filter or produce a syntax error. Instead, it
changes the observable behavior of the stored procedure by placing report output
inside a path that is skipped when the cursor is empty. In procedural database
programming, \texttt{DBMS\_OUTPUT} calls are part of the program's external
behavior, and exception handling must preserve the required side effects. This
example shows that even compiling PL/SQL can fail when models mishandle the
ordering of cursor iteration, output emission, and exception control flow.
\end{tcolorbox}

\captionof{figure}{Representative wrong-procedure-logic failure. The generated
procedure compiles, but raises \texttt{NO\_DATA\_FOUND} before emitting the
required report headers, so the observable \texttt{DBMS\_OUTPUT} behavior does
not match the reference.}
\label{fig:error-logic-output}
\noindent
\begin{tcolorbox}[errorcase]
\textbf{Error Category: Invalid or incomplete PL/SQL artifact.}

\textbf{Example.}
Dataset: \texttt{Spider};
Model: \texttt{GPT-5.4};
Instance: \texttt{spider\_237};
Tests passed: \texttt{0/1}.

\textbf{Task.}
The user asks for PL/SQL code that prints the titles of all cartoons on the TV
channel whose series name is \texttt{Sky Radio}, with no output header. The task
is solvable from the provided schema by joining \texttt{TVSHOW.TV\_CHANNEL} with
\texttt{TVSHOW.CARTOON} and printing the matching titles.

\textbf{Failure signal.}
Instead of returning an executable PL/SQL block, the model outputs:

\errline{INVALID\_REQUEST}

The expected output contains two titles:

\errline{The Rise of the Blue Beetle!}
\errline{Return of the Fearsome Fangs!}

The actual output is empty. Evaluation fails before semantic comparison because
the harness attempts to execute \texttt{INVALID\_REQUEST} as SQL:

\errline{ORA-00900: invalid SQL statement}

\textbf{Why this is an error.}
This is an artifact-construction failure. The model rejects or fails to
instantiate a valid task and returns text that is not executable PL/SQL. The
problem is not an incorrect filter or output row inside executable code, but the
absence of a valid database artifact.
\end{tcolorbox}
\captionof{figure}{Representative invalid-artifact failure. The model returns
\texttt{INVALID\_REQUEST} for a solvable query, causing execution to fail before
semantic comparison.}
\label{fig:error-invalid-artifact}

\vspace{0.75em}

\noindent
\begin{tcolorbox}[errorcase]
\textbf{Error Category: Interface, signature, or schema grounding.}

\textbf{Example.}
Dataset: \texttt{Spider2-MT-dev};
Model: \texttt{Gemini-2.5-Flash-Lite};
Instance: \texttt{plsql-multiturn-0739\_q04};
Tests passed: \texttt{0/4}.

\textbf{Task.}
The user asks for the package procedure
\texttt{PIZZA\_RUNNER\_UTIL.KPI\_DASHBOARD}, which accepts a runner ID and
prints that runner's delivery KPIs.

\textbf{Failure signal.}
The generated code defines the procedure only inside:

\errline{CREATE OR REPLACE PACKAGE BODY MODERN\_DATA.PIZZA\_RUNNER\_UTIL}

It does not create the required public package specification at the expected
entry point. The harness invokes:

\errline{BEGIN pizza\_runner\_util.kpi\_dashboard(1); END;}

All four tests therefore fail before output comparison with:

\errline{PLS-00201: identifier 'PIZZA\_RUNNER\_UTIL.KPI\_DASHBOARD' must be declared}

\textbf{Why this is an error.}
This is a contract-preservation failure. Although the generated package body
contains plausible KPI logic, schema-qualifying the package and omitting its
callable specification means that the required public procedure cannot be
resolved by the benchmark harness.
\end{tcolorbox}

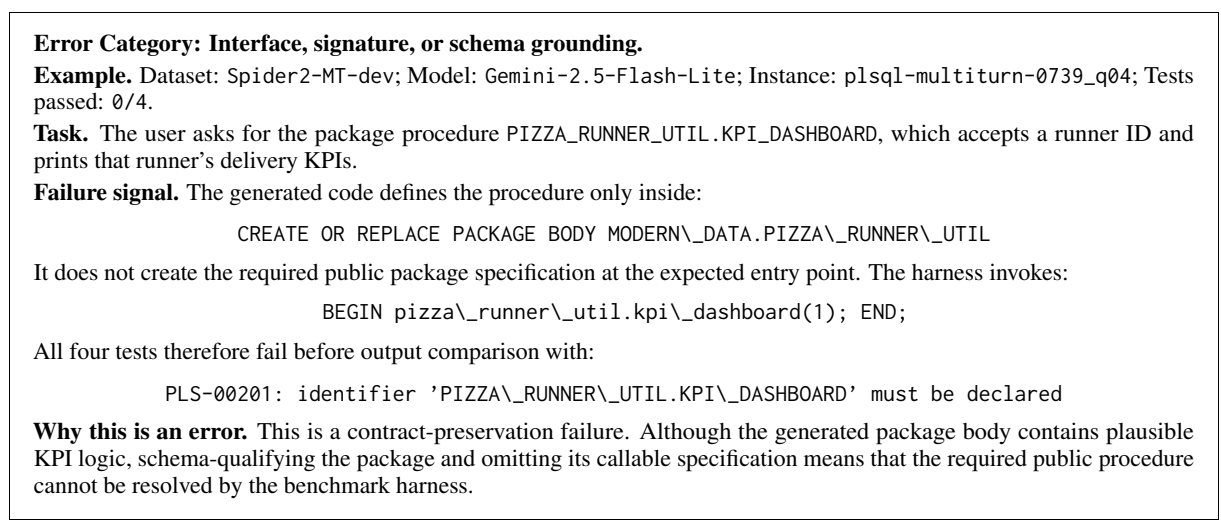
\captionof{figure}{Representative interface failure. The generated artifact
creates only a schema-qualified package body, so the required public package
procedure cannot be called.}
\label{fig:error-interface-schema}

% Use this only if there is more two-column content after the appendix examples.
% \clearpage
% \twocolumn

% System prompt
% \onecolumn
\section{Prompts}
\label{app:prompts}
\subsection{Prompt for PL/SQL Code Generation}
We followed the SQL generation system prompt structure from \citep{liu2025oraplan,somayajula2026soma}:
\begin{figure*}[!htbp]
\small
\centering
\begin{tcolorbox}[colback=white,colframe=black,width=\textwidth,
                  boxrule=0.5pt,arc=0pt,outer arc=0pt,
                  left=6pt,right=6pt,top=6pt,bottom=6pt]

You are an expert database developer proficient in Oracle PL/SQL.

\textbf{Goal:} Generate executable Oracle PL/SQL that correctly satisfies the user request using only the provided schema context.

\textbf{Background:} PL/SQL is Oracle’s procedural extension to SQL. It supports
variables, control flow, exception handling, cursors, packages, object
types, and dynamic SQL. Correct PL/SQL must be schema-grounded,
Oracle-dialect compliant, and executable.

\textbf{Input}
\begin{itemize}
    \item \textbf{DDL context:} \texttt{\{ddl\}}
    \item \textbf{User request:} \texttt{\{question\}}
\end{itemize}

\textbf{Instructions:}
\begin{itemize}
    \item Use only tables, columns, and database objects that appear in the DDL context.
    \item Prefer fully qualified names in \texttt{SCHEMA.TABLE} format when available.
    \item Generate Oracle-dialect PL/SQL only.
    \item Output the requested artifact type, such as an anonymous block, procedure,
function, package, package body, or a set of related CREATE OR REPLACE
statements as needed.
    \item Include appropriate exception handling when the task involves data lookup or iterative processing.
    \item Do not include explanations, comments outside the code, or markdown.
\end{itemize}

\textbf{Output Format:}
\begin{itemize}
    \item Return only executable PL/SQL text.
    \item Include the trailing \texttt{/} on a new line.
    \item If the request cannot be satisfied because required tables, columns, or database objects are missing, or because the task is logically impossible, return exactly: INVALID\_REQUEST
\end{itemize}

\end{tcolorbox}
\caption{Prompt for PL/SQL generation.}
\label{fig:plsql_generation_prompt}
\end{figure*}

%hyperparam setting
\newpage

\section{Hyperparameters and Infrastructure}
\label{app:hyperparams}

To support reproducibility, we report the model, prompting, inference, and infrastructure settings used in our \textsc{PLSQLBench} experiments.

\begin{table*}[!htbp]
  \centering
  \footnotesize
  \setlength{\tabcolsep}{3pt}
  \renewcommand{\arraystretch}{1.03}
  \begin{tabularx}{0.9\textwidth}{llX}
    \toprule
    \textbf{Module} & \textbf{Hyperparameter} & \textbf{Value / Notes} \\
    \midrule
Models & Evaluated models & Gemma-4-31B, Llama-4-Maverick, Claude-Opus-4.8, Gemini-2.5-Flash-Lite, Grok-4.3, GPT-5.4-Mini, GPT-5.4, and GPT-5.6-Sol \\
          % & Backend & OCI Generative AI / model provider endpoints \\
    \midrule
    Prompts & System prompt & Oracle PL/SQL program-unit generation prompt, see Appendix \ref{app:prompts} \\
            & Prompt inputs & Natural-language request, schema DDL context, and task requirements \\
            & Multi-turn context & Previous turn question and generated answer appended to subsequent turns \\
            & Output format & Executable Oracle PL/SQL only; trailing \texttt{/}; no markdown explanations \\
    \midrule
    Inference & Temperature & 0.0 \\
              & Max. output tokens & 8192 \\
              & Context mode & Rollout context: each turn is generated using the previous turns' model-generated responses as conversation history \\
              & Regeneration policy & Failed, empty, and retry-eligible cached generations regenerated \\
    \midrule
    Infrastructure & Database & Oracle Autonomous Database 23ai \\
                   & Database service & Serverless Autonomous Database \\
                   & Execution environment & Oracle PL/SQL execution over benchmark schemas and test cases \\
                   & Database documentation & \url{https://docs.oracle.com/en/cloud/paas/autonomous-database/serverless/adbsb/autonomous-always-free-23ai.html} \\
    \bottomrule
  \end{tabularx}
  \caption{Hyperparameter and infrastructure settings for \textsc{PLSQLBench}.}
  \label{tab:plsqlbench_hyperparams}
\end{table*}

\end{document}